\documentclass[runningheads]{llncs}

\usepackage{amsfonts}
\usepackage{graphicx}
\usepackage{multirow}
\usepackage{cite}
\usepackage{subfigure}
\usepackage{booktabs}
\usepackage{url}
\usepackage{floatrow}
\usepackage{afterpage}

\begin{document}

\title{CELNet: Evidence Localization for Pathology Images using Weakly Supervised Learning}
\titlerunning{Evidence Localization for Pathology Images}

\author{Yongxiang Huang \and Albert C. S. Chung}

\authorrunning{Y. Huang and A. C. S. Chung}

\institute{
	Lo Kwee-Seong Medical Image Analysis Laboratory,\\
	Department of Computer Science and Engineering,\\
	The Hong Kong University of Science and Technology, Hong Kong, China\\
	\email{\{yhuangch,achung\}@cse.ust.hk}
}
\maketitle

\begin{abstract}
Despite deep convolutional neural networks boost the performance of image classification and segmentation in digital pathology analysis, they are usually weak in interpretability for clinical applications or require heavy annotations to achieve object localization. To overcome this problem, we propose a weakly supervised learning-based approach that can effectively learn to localize the discriminative evidence for a diagnostic label from weakly labeled training data. Experimental results show that our proposed method can reliably pinpoint the location of cancerous evidence supporting the decision of interest, while still achieving a competitive performance on glimpse-level and slide-level histopathologic cancer detection tasks. 

\end{abstract}

\section{Introduction}

\label{section1}

Pathology analysis based on microscopic images is a critical task in medical image computing. In recent years, deep learning of digitalized pathology slide has facilitated the progress of automating many diagnostic tasks, offering the potential to increase accuracy and improve review efficiency. Limited by computation resources, deep learning-based approaches on whole slide pathology images (WSIs) usually train convolutional neural networks (CNNs) on patches extracted from WSIs and aggregate the patch-level predictions to obtain a slide-level representation, which is further used to identify cancer metastases and stage cancer \cite{Wang2016DeepLF}. Such a patch-based CNN approach has been shown to surpass pathologists in various diagnostic tasks \cite{Liu2017DetectingCM} 

Off-the-shelf CNNs have been shown to be able to accurately classify or segment pathology images into different diagnostic types in recent studies \cite{huang2018improving, veeling2018rotation}. However, 
most of these methods are weak in interpretability especially for clinicians, due to a lack of evidence supporting for the decision of interest. During diagnosis, a pathologist often inspects abnormal structures (e.g., large nucleus, hypercellularity) as the evidence for determining whether the glimpsed patch is cancerous. For CAD systems, learning to pinpoint the discriminative evidence can provide precise visual assistance for clinicians. Strong supervision-based feature localization methods require a large number of pathology images annotated in pixel-level or object-level, which are very costly and time-consuming and can be biased by the experiences of the observers. In this paper, we propose a weakly supervised learning (WSL) method that can learn to localize the discriminative evidence for the class-of-interest on pathology images from weakly labeled (i.e. image-level) training data. Our contributions include: 
i) proposing a new CNN architecture with multi-branch attention modules and deep supervision mechanism, to address the difficulty of localizing discrete and small objects in pathology images, 
ii) formulating a generalizable approach that leverages gradient-weighted class activation map and saliency map in a complementary way to provide accurate evidence localization, 
iii) designing a new attention module which allows capturing spatial attention from various context, 
iv) quantitatively and visually evaluating WSL methods on large scale histopathology datasets, and 
v) constructing a new dataset (HPLOC) based on Camelyon16 for effectively evaluating evidence localization performance on histopathology images.


\textbf{Related Work.} Recent studies have demonstrated that CNN can learn to localize discriminative features even when it is trained on image-level annotations \cite{zhou2016learning}. However, these methods are evaluated on natural image datasets (e.g., PASCAL), where the objects of interest are usually large and distinct in color and shape. In contrast, objects in pathology images are usually small and less distinct in morphology between different classes. A few recent studies investigated WSL approaches on medical images, including lung nodule detection and placental ultrasound images \cite{feng2017discriminative}. These methods employ GAP-based class activation map and require CNNs ending with global average pooling, which degrades the performance of CNNs as a side effect \cite{zhou2016learning}. 

\section{Methods}
The overview of the framework is shown in Fig.{~\ref{fig:fig1}}. 
The model is trained to predict the cancer score for a given image, indicating the presence of cancer metastasis. 
In the test phase, besides giving a binary classification, the model generates a cancerous evidence localization map and performs localization. 

 \afterpage{
 \begin{figure}[t]
 	\centering

 	\begin{tabular}{cc}
 		\includegraphics[scale=0.5]{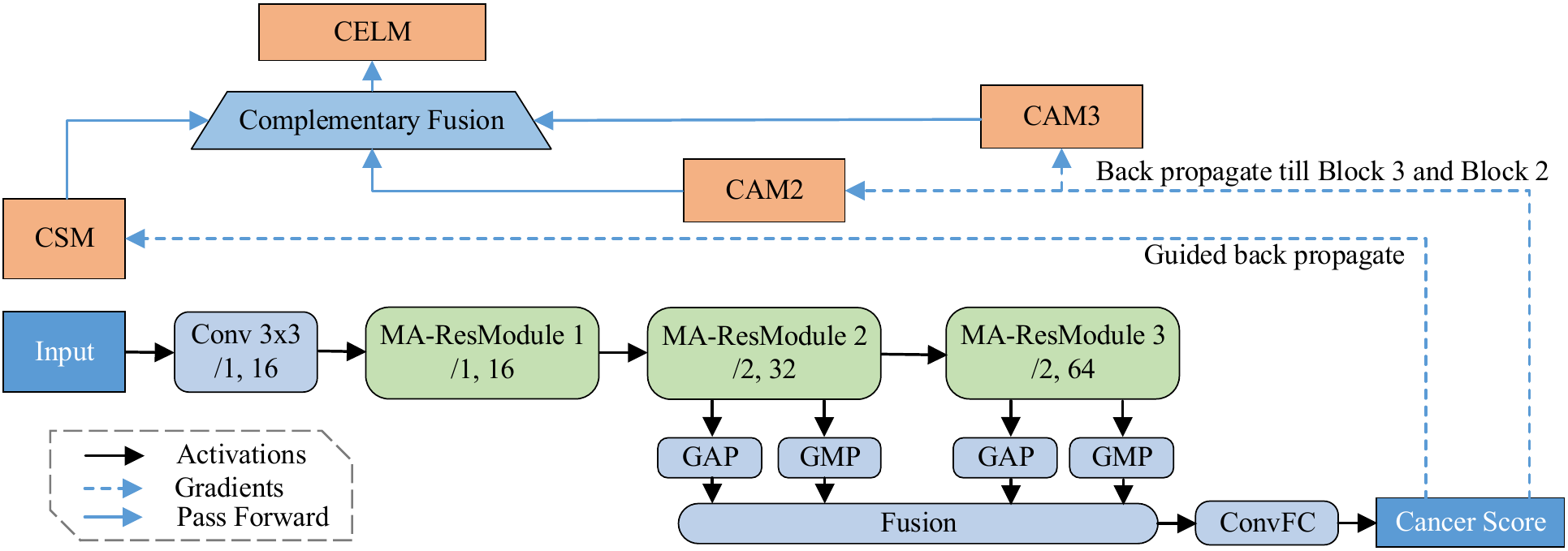} & 
 

 		\includegraphics[scale=0.5]{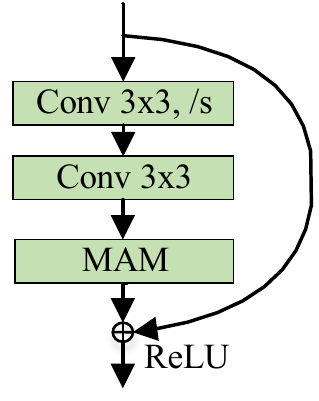} \\

 	\end{tabular}
 	\caption{ \textbf{Left:} Framework overview of the proposed WSL method. The under line (e.g., /1, 16) denotes stride and number of channels.  \textbf{Right:} A building block for the multi-branch attention based residual module (MA-ResModule). 
 	} 
 	\label{fig:fig1}
 \end{figure}
}
 
 \subsection{Cancerous Evidence Localization Networks (CELNet)}
 \label{subsection_2_3}

 Given the object of interest is relatively small and discrete, a moderate number of convolutional layers is sufficient for encoding locally discriminative features. As discussed in Section \ref{section1}, instances on pathology images are similar in morphology and can be densely distributed, the model should avoid over-downsampling in order to pinpoint the cancerous evidence from the densely distributed instances. The proposed CELNet starts with a $3\times 3$ convolution head followed by 3 Multi-branch Attention-based Residual Modules (MA-ResModule)
 \footnote{Densely connected module is not employed considering it is comparatively speed-inefficient for WSIs application due to its dense tensor concatenation. } 
 \cite{he2016deep}. Each MA-ResModule is composed of 3 consecutive building blocks integrated with the proposed attention module (MAM) as  shown in Fig.{~\ref{fig:fig1}} (Right). We use $3\times 3$ convolution with stride of 2 for downsampling in residual connections instead of $1\times 1$ convolution to reduce information loss. Batch normalization and ReLU are applied after each convolution layer for regularization and non-linearity.
 
 \subsubsection{Multi-branch Attention Module (MAM)}
 To eliminate the effect of background contents and focus on representing the cancerous evidence (which can be sparse), we employ attention mechanism. 
 Improved on Convolutional Block Attention Module (CBAM) 
 , which extracts channel attention and spatial attention of an input feature map in a squeeze and excitation manner, we propose a multi-branch attention module. MAM can better approximate the importance of each location on the feature map by looking at its context at different scales. 
 Given a squeezed feature map $F_{sq} $ generated by the channel attention module, we compute and derive a 2D spatial attention map $A_s$ by $ A_s = \sigma(\sum_{k'} f^{k' \times k'} (F_{sq}) ),$
 where $f^{k' \times k'}$ represents a convolution operation with kernel size of $k' \times k'$, and $\sigma$ denotes the sigmoid function. We set $k' \in \{3, 5, 7\}$ in our experiments, corresponding to 3 branches. Hereby, the feature map  $F_{sq} $ is refined by element-wise multiplication with the spatial attention map $A_s$.
 
MAM is conceptually simple but effective in improving detection and localization performance as demonstrated in our experiments. 

 \subsubsection{Deep Supervision}
 Deep supervision \cite{lee2015deeply} is employed to empower the intermediate layers to learn class-discriminative representations, for building the cancer activation map in a higher resolution.  We achieve this by adding two companion output layers to the last two MA-ResModules, as shown in Fig. \ref{fig:fig1}. Global max pooling (GMP) is applied to search for the best discriminative features spatially, while global average pooling (GAP) is applied to encourage the network to identify all discriminative parts on the image. Each companion output layer applies GAP and GMP on the input feature map and concatenates the resulting vectors. The cancer score of the input image is derived by concatenating the outputs of the two companion layers followed by a fully convolutional layer (i.e., kernel size $1 \times 1$) with a sigmoid activation. 
 CELNet enjoys efficient inference when applied to test WSIs, as it is fully convolutional and avoids repetitive computation for the overlapping part between neighboring patches.

\subsection{Cancerous Evidence Localization Map (CELM) }
\subsubsection{Cancer Activation Map (CAM)}
Given an image $I \in \mathbb{R}^{H \times W \times 3}$, let $y^c = S_c(I)$ represent the cancer score function governed by the trained CELNet (before sigmoid layer). 
A cancer-class activation map $M^c$ shows the importance of each region on the image to the diagnostic value. For a target layer $l$, the CAM $M^c_l$ is derived by taking the weighted sum of feature maps $F_l$  with the weights \{$\alpha_{k,l}^c$ \}, where $\alpha_{k,l}^c$ represents the importance of $k^{th}$ feature plane. The weights $\alpha_{k,l}^c$ are computed as $\alpha_{k,l}^c = Avg_{i,j}( \frac{\partial y^c}{\partial F_l^k(i,j)} )$, i.e., spatially averaging the gradients of  cancer score $y^c$ with respect to the $k^{th}$ feature plane $F_l^k$, which is achieved by back propagation (see Fig.\ref{fig:fig1}). Thus, the CAM of layer $l$ can be derived by $ M_l^c = ReLU(\sum_k \alpha_{k,l}^c F_l^k)$, where ReLU is applied to exclude the features with negative influence on the class of interest \cite{selvaraju2017grad}. 

We derive two CAMs, $M^c_2$ and $M^c_3$ from the last layer of the second and the third residual module on CELNet respectively (i.e., CAM2 and CAM3 in Fig.\ref{fig:fig1}). CAM3 can represent discriminative regions for identifying a cancer class  in a relatively low resolution while CAM2 enjoys higher resolution and still class-discriminative under deep supervision. 

\subsubsection{Cancer Saliency Map (CSM)}
In contrast with CAM, the cancer-class saliency map shows the contribution of each pixel site to the cancer score $y^c$. This can be approximated by the derivate of a linear function $S^c(I) \approx w^TI + b$. Thus the pixel contribution is computed as $ w = \frac{\partial S^c(I)}{\partial I} $. Different from \cite{Simonyan2013DeepIC}, we derive $w$  by the guided back-propagation \cite{springenberg2014striving} to prevent backward flow of negative gradients. 
For a RGB image, to obtain its cancer saliency map $M^s \in \mathbb{R}^{H \times W\times 1} $ from $w  \in \mathbb{R}^{H \times W \times 3} $, we first normalize $w$ to $[0,1]$ range, followed by greyscale conversion and Gaussian smoothing, instead of simply taking the maximum magnitude of $w$ as proposed in \cite{Simonyan2013DeepIC}. Thus, the resulting cancer saliency map (see Fig.{~\ref{fig:vis}} (b)) is far less noisy and more focus on class-related objects than the original one proposed in \cite{Simonyan2013DeepIC}. 

\subsubsection{Complementary Fusion}
The generated CAMs coarsely display discriminative regions for identifying a cancer class (see Fig.\ref{fig:vis} (c)), while the CSM is fine-grained, sensitive and represents pixelated contributions for the identification (see Fig.\ref{fig:vis} (b)). To combine the merits of them for precise cancerous evidence localization, we propose a complementary fusion method. First, CAM3 and CAM2 are combined to obtain a unified cancer activation map $M^c \in \mathbb{R}^{H \times W\times 1}$   as $M^c = \alpha f_u(M^c_3) + (1- \alpha) f_u(M^c_2)$, where $f_u$ denotes a upsampling function by bilinear interpolation, and  the coefficient $\alpha$ in range [0,1] is confirmed by validation. 
The CELM is derived by complementarily fusing CSM and CAM as $M = \beta (M^c \odot M^s) + (1 - \beta) M^c$,
where $\odot$ denotes element-wise product, and  the coefficient $\beta$ captures the reliability of the point-wise multiplication of CAM and CSM, and the value of $\beta$ is estimated by cross-validation in experiments.

\section{Experiments \& Results}
We first evaluate the detection performance of the proposed model as for clinical requirements, followed by evidence localization evaluations.   

\subsection{Datasets and Experimental Setup}
The detection performance of the proposed method is validated on two benchmark datasets, PCam\cite{veeling2018rotation} and Camelyon16 \footnote{https://camelyon16.grand-challenge.org}. 

\textbf{PCam:} The PCam dataset contains 327,680 lymph node histopathology images of size $96 \times 96$ with binary class labels indicating the presence of cancer metastasis, split into 75\% for training, 12.5\% for validation, and 12.5\% for testing as originally proposed. The class distribution in each split is balanced (1:1). For a fair comparison, following \cite{veeling2018rotation}, we perform image augmentation by random 90-degree rotations and horizontal flipping during training. 

\textbf{Camelyon16:} The Camelyon16 dataset includes 270 H\&E stained WSIs (160 normal and 110 cancerous cases) for training and 129 WSIs held out for testing (80 normal and 49 cancerous cases) with average image size about $65000 \times 45000$, where regions with cancer metastasis are delineated in cancerous slides. To apply our CELNet on WSIs, we follow the pipeline proposed in \cite{Liu2017DetectingCM}, including WSI pre-processing, patch sampling and augmentation, heatmap generation, and slide-level detection tasks. For slide-level classification, we take the maximum tumor score among all patches as the final slide-level prediction. For tumor region localization, we apply non-suppression maximum algorithm on the tumor probability map aggregated from patch predictions to iteratively extract tumor region coordinates. We work on the WSI data at 10$\times$ resolution instead of  40$\times$ with the available computation resources. 

	In our experiments, all models are trained using binary cross-entropy loss with L2 regularization of $10^{-5}$ to improve model generalizability, and optimized by SGD with Nesterov momentum of 0.9 with a batch size of 64 for 100 epochs. The learning rate is initialized with $10^{-4}$ and is halved at 50 and 75 epochs. We select model weights with minimum validation loss for test evaluation.   

\subsection{Classification Results}	

As Tbl.{~\ref{table1}} shows, CELNet consistently outperforms ResNet, DenseNet, and P4M-DenseNet \cite{veeling2018rotation} in histopathologic cancer detection on the PCam dataset. 
 P4M-DenseNet uses less parameters due to parameter sharing in the p4m-equivariance. 
 For auxiliary experiments, we perform ablation studies and visual analysis. From Tbl.{~\ref{table1}} , we observe that our attention module brings 1.77\% accuracy gain, which is larger than the gain brought by CBAM \cite{woo2018cbam}. Both the CAM and CELM on CELNet are mainly activated for the cancerous regions (see Fig.\ref{fig:vis} (c) and (d)). These subfigures indicate that CELNet is effective in extracting discriminative evidence for histopathologic classification.  
 
\begin{table}[h]
	\centering
	\floatbox[{\capbeside\thisfloatsetup{capbesideposition={left,top} }}]{table}[\FBwidth]
		{
		\caption{Quantitative comparisons on the PCam test set. P4M-DenseNet  \cite{veeling2018rotation}:  current SoTA method for  the PCam benchmark, CELNet: our method, $^{-}$: removal of the proposed multi-branch attention module, +CBAM: integration with convolutional block attention module \cite{woo2018cbam}.  
		}}
		{
			\begin{tabular}{l  c c c }
				\toprule
				Methods							&   Acc 		& AUC 	& \#Params \\
				\midrule    
				ResNet18 \cite{he2016deep}			& 	  88.73			&	95.36		& 11.2M	\\
				DenseNet \cite{veeling2018rotation}				&	 87.20  & 94.60	& 902K   \\
				P4M-DenseNet    & 89.80 	&	96.30	& 119K \\
				\midrule  
				CELNet		  			    &	 \textbf{91.87}	& \textbf{97.72} & 297K	 \\
				CELNet$^{-}$ 	 			&	90.10 		& 96.45	& 292K		\\
				CELNet$^{-}$ +CBAM &	  90.86 	& 97.17 & 	  296K	\\
				\bottomrule
			\end{tabular}
		\label{table1}
		}
\end{table}

On slide-level detection tasks, as shown in Tbl.\ref{table2}, our CELNet based approach achieves higher classification performance (1.7\%) in terms of AUC than the baseline method \cite{Liu2017DetectingCM}, and  outperforms previous state-of-the-art methods in slide-level tumor localization performance in terms of FROC score. The results illustrate that instead of using off-the-shelve CNNs as the core patch-level model for histopathologic slide detection, adopting CELNet can potentially bring larger performance gain. CELNet is more parameter-efficient as shown in Tbl.\ref{table1} and testing a slide on Camelyon16 takes about 2 minutes on a Nvidia 1080Ti GPU. 
\begin{table}[h]
	\centering
	\floatbox[{\capbeside\thisfloatsetup{capbesideposition={left,top} }}]{table}[\FBwidth]
	{\caption{Quantitative comparisons of slide-level classification performance (AUC) and slide-level tumor localization performance (FROC) on the Camelyon16 test set. *: The Challenge Winner uses $40\times$ resolution while results of other methods are based on $10\times$. 
	}}
	{
		\begin{tabular}{l  c c  }
			\toprule
			Methods							&   AUC 		& FROC 	 \\
			\midrule    
			P4M-DenseNet			& 	  -												&	84.0 			\\
			Liu \cite{Liu2017DetectingCM}    & 96.5 						&	79.3  	 \\
			Challenge Winner$^*$ \cite{Wang2016DeepLF}    & \textbf{99.4} 	&	80.7	 \\
			Pathologist 				&	  			96.6 							& 73.3	\\
			CELNet		  			    &	  97.2 & \textbf{84.8} 	 \\
			\bottomrule
		\end{tabular}
		\label{table2}
	}
\end{table}

\subsection{Weakly Supervised Localization and Results}
Given that the trained CELNet can precisely classify a pathology image, here we aim to investigate its performance in localizing the supporting evidence based on the proposed CELM. To achieve this, based on Camelyon16, we first construct a dataset with region-level annotations for cancer metastasis, namely HPLOC, 
and develop the metrics for measuring localization performance on HPLOC. 

\textbf{HPLOC:} The HPLOC dataset contains 20,000 images of size $96 \times 96$ with segmentation masks for cancerous region. Each image is sampled from the test set of Camelyon16 and contains both cancerous regions and normal tissue in the glimpse, which harbors the high quality of the Camelyon16 dataset. 

\begin{figure}[t]
	\centering

	\begin{tabular}{cccccc}


		\includegraphics[scale=0.5]{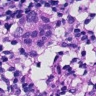}  & 
		\includegraphics[scale=0.5]{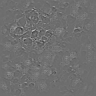} &
		\includegraphics[scale=0.5]{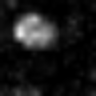} &
		\includegraphics[scale=0.5]{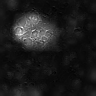} & 

		\includegraphics[scale=0.5]{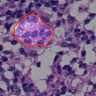} &
		\includegraphics[scale=0.5]{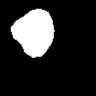}
 \\
		(a) Input  & (b) CSM  & (c) CAM & (d) CELM  & (e) Localization  & (f) GT   \\
	\end{tabular}

	\caption{ Evidence localization results of our WSL method on the HPLOC dataset.
		 (a) Input glimpse, (b) Cancer Saliency Map, (c) Cancer Activation Map, (d) CELM: Cancerous Evidence Localization Map, (e) Localization results based on CELM,  where the localized evidence is highlighted for providing visual assistance, (f) GT: ground truth, white masks represent tumor regions and the black represents normal tissue} 

	\label{fig:vis}

\end{figure}

\textbf{Metrics:} To perform localization, we generate segmentation masks from CELM/CAM/CSM by thresholding and smoothing (see Fig.\ref{fig:vis} (e)). If a segmentation mask intersects with the cancerous region by at least 75\%
\footnote{The annotated contour in Camelyon16 is usually enlarged to surround all tumors. }
, it is defined as a true positive. Otherwise, if a segmentation mask intersects with the normal region by at least 75\%, it is considered as a false positive. Thus, we can use precision and recall score to quantitatively assess the localization performance of different WSL methods, where the results are summarized in Tbl.\ref{table3}. 

\begin{table}[h]
	\centering
	\floatbox[{\capbeside\thisfloatsetup{capbesideposition={left,top} }}]{table}[\FBwidth]
	{\caption{Quantitative comparisons for different weakly supervised localization methods on the HPLOC dataset. Ours: CELNet + CELM. MAM and DS are short for multi-branch attention module and deep supervision respectively. 
	}}
	{
		\begin{tabular}{l  c c }
			\toprule
			Methods				& Precision & Recall \\
			\midrule   
			ResNet18 + Backprop \cite{Simonyan2013DeepIC}     		& 79.8	& 85.5 \\  
			ResNet18 + GradCAM \cite{selvaraju2017grad}		  	& 	 85.6	& 82.4   \\
			\midrule  
			Ours		  		&	 \textbf{91.6}	& 87.3  	 \\
			Ours w/o MAM	 		&	88.1 				& 85.6	\\
			Ours w/o DS		&	  90.5 		& 	\textbf{87.7}	\\
			CELNet + GradCAM  &	  91.0		& 	85.4	\\
			\bottomrule
		\end{tabular}
		\label{table3}
	}
\end{table}

We observe that our WSL method based on CELNet and CELM consistently performs better than the back propagation-based approach \cite{Simonyan2013DeepIC} and the class activation map-based approach \cite{selvaraju2017grad}. Note that we used ResNet18 \cite{he2016deep} as the backbone for the compared methods because it achieves better classification performance and provides higher resolution for GradCAM (12$\times$12) as compared to DenseNet (3 $\times$ 3) \cite{veeling2018rotation}. 
We perform ablation studies to further evaluate the key components of our method in Tbl.\ref{table3}. We observe the effectiveness of the proposed multi-branch attention module in increasing the localization accuracy. 
The deep supervision mechanism effectively improves the precision in localization despite slightly lower recall score, which can be caused by the regularization effect on the intermediate layers, that is, encouraging the learning of discriminative features for classification but also potentially discouraging the learning of some low-level histological patterns. 
We observe that using CELM can improve the recall score and precision, which indicates that CELM allows better discovery of cancerous evidence than using GradCAM. 
We present the visualization results in Fig. {\ref{fig:vis}}, the cancerous evidence 
is represented as large nucleus and hypercellularity in the images, which are precisely captured by the CELM. Fig.\ref{fig:vis}(e) visualizes the localization results by overlaying the segmentation mask generated from CELM onto the input image, which demonstrates the effectiveness of our WSL method in localizing cancerous evidence. 

\section{Discussion \& Conclusions}
In this paper, we have proposed a generalizable method for localizing cancerous evidence on histopathology images. 
Unlike the conventional feature-based approaches, the proposed method does not rely on specific feature descriptors but learn discriminative features for localization from the data. 
To the best of our knowledge, investigating weakly supervised CNNs for cancerous evidence localization and quantitatively evaluating them on large datasets have not been performed on histopathology images. 
Experimental results show that our proposed method can achieve competitive classification performance on histopathologic cancer detection, and more importantly, provide reliable and accurate cancerous evidence localization using weakly training data, which reduces the burden of annotations. We believe that such an extendable method can have a great impact in detection-based studies in microscopy images and help improve the accuracy and interpretability for current deep learning-based pathology analysis systems. 

\bibliographystyle{splncs04}

\bibliography{main.bib}

\end{document}